\documentclass{article}

\PassOptionsToPackage{numbers}{natbib}

     \usepackage[preprint]{neurips_2022}

\usepackage[utf8]{inputenc} 
\usepackage[T1]{fontenc}    
\usepackage{hyperref}       
\usepackage{url}            
\usepackage{booktabs}       
\usepackage{amsfonts}       
\usepackage{nicefrac}       
\usepackage{microtype}      
\usepackage{xcolor}         

\usepackage{graphicx,wrapfig}

\usepackage{amsmath}
\usepackage{amssymb}
\usepackage{mathtools}
\usepackage{amsthm}
\usepackage{amsthm}
\usepackage{mathrsfs}
\usepackage{stmaryrd}
\usepackage{caption}
\usepackage{subcaption}
\usepackage{multirow}
\usepackage{paralist}
\usepackage{enumitem}
\usepackage{algorithm}
\usepackage{algorithmic}
\usepackage{arydshln}

\usepackage{./notations}

\usepackage[capitalize,noabbrev]{cleveref}

\theoremstyle{plain}

\theoremstyle{definition}

\theoremstyle{remark}

\title{Algorithms for Weak Optimal Transport with an Application to Economics}

\author{%
  Fran\c{c}ois-Pierre Paty\\
  CREST-ENSAE\\
  Institut Polytechique de Paris\\
  \texttt{francois.pierre.paty@ensae.fr} \\
  \And
  Philippe Chon\'e\\
  CREST-ENSAE\\
  Institut Polytechique de Paris\\
  \texttt{philippe.chone@ensae.fr} \\
  \And
  Francis Kramarz\\
  CREST-ENSAE\\
  Institut Polytechique de Paris\\
  \texttt{francis.kramarz@ensae.fr}
}

\begin{document}

\maketitle

\begin{abstract}


The theory of weak optimal transport (WOT), introduced by~\cite{Gozl_Robe_Teta_JFA_2017}, generalizes the classic Monge-Kantorovich framework by allowing the transport cost between one point and the points it is matched with to be nonlinear.
In the so-called barycentric version of WOT, the cost for transporting a point $x$ only depends on $x$ and on the barycenter of the points it is matched with.
This aggregation property of WOT is appealing in  machine learning, economics and finance.
Yet algorithms to compute WOT have only been developed for the special case of quadratic barycentric WOT, or depend on neural networks with no guarantee on the computed value and matching. The main difficulty lies in the transportation constraints which are costly to project onto. In this paper, we propose to use mirror descent algorithms to solve the primal and dual versions of the WOT problem.
We also apply our algorithms to the variant of WOT introduced by~\cite{chogozkra2021} where mass is distributed from one space to another through unnormalized kernels (WOTUK). 
We empirically compare the solutions of WOT and WOTUK with classical OT. We illustrate our numerical methods to the economic framework of~\cite{chone2021matching}, namely the matching between workers and firms on labor markets. 

\end{abstract}
\section{Introduction}

Over the past few years, optimal transport (OT) has gained importance in the machine learning community as a useful tool to analyze data, with applications to various domains such as graphics~\citep{2015-solomon-siggraph,2016-bonneel-barycoord}, imaging~\citep{rabin2015convex,2016-Cuturi-siims}, generative models~\citep{WassersteinGAN,salimans2018improving}, biology~\cite{hashimoto2016learning,schiebinger2019optimal}, NLP~\cite{grave2018unsupervised,alaux2018unsupervised}, finance~\cite{beiglbock2013model, GalichonMartingale, acciaio2020weak} or economics~\cite{galichon2015cupid, galichon2016optimal, lindenlaub2017sorting}.

The key in making the optimal transport approach work in these applications lies in the different forms of regularization added to the classical optimal transport problem. Although many different types of regularizations have been proposed in the literature, among which the celebrated entropic regularization~\cite{CuturiSinkhorn}, most of these correspond to penalized versions of the Monge-Kantorovich problem, possibly with relaxed~\cite{FigalliPartial, chizat2017unbalanced} or tightened~\cite{beiglbock2013model, paty-SSNB} constraints.

\looseness=-1 These variants of OT are unable to capture the aggregation property that plays a crucial role in some applications~\citep{backhoff2020applications, chone2021matching}. In these applications, the cost for matching points $x$ and $y$ does not only depends on $x$ and $y$ through a cost function, but also on all the other points $y'$ matched with $x$. This idea was formalized by~\citep{Gozl_Robe_Teta_JFA_2017, alibert:hal-01741688} and gave rise to the notion of \textit{weak optimal transport}, which generalizes OT. Even more recently, and guided by the economic application of~\cite{chone2021matching}, \cite{chogozkra2021}~proposed to relax the WOT problem and defined the notion of \textit{weak optimal transport with unnormalized kernel}, henceforth WOTUK.

In the discrete setting, when the cost function is convex, weak optimal transport (WOT) defines a convex optimization problem over the transportation polytope~\cite{pmlr-v119-paty20a}. Nevertheless, algorithms to compute WOT when the measures are discrete have only been proposed in the special case of quadratic barycentric WOT~\cite{cazelles2021streaming} (see subsection~\ref{par:barycentric_wot} for a precise definition). In their recent preprint, \cite{korotin2022neural}~propose to use neural networks to approximate the WOT problem, but do not not provide guarantees for their optimization procedure.

\paragraph{Contributions}
The goal of this paper is twofold: first, we present recent optimal transport problems of interest in economics, the so-called WOT and WOTUK problems, to the machine learning community; second, we introduce algorithms to solve these problems numerically, when the considered measures are discrete. We compare these variants of OT with classical OT and apply our algorithms to the economics framework of~\cite{chone2021matching}.

\paragraph{Structure of the paper}
We begin in section~\ref{sec:wot} by reminders on optimal transport and weak optimal transport problems, which we introduce in the context of the economic application of~\cite{chone2021matching}. We continue with the definition and the key properties of weak optimal transport with unnormalized kernel in section~\ref{sec:wotuk}. We present our algorithms in section~\ref{sec:algo}, and conclude the paper in section~\ref{sec:expes} with numerical experiments and applications in economics.
\section{From Optimal Transport to Weak Optimal Transport} \label{sec:wot}

\subsection{Optimal Transport and Matching}

A classic problem in labor economics~\cite{Kels_Craw_Ecma_1982,Heck_Sche_ReStud_1987,Eeck_Kirc_Eca_2018} is to understand the matching between workers and firms, \textit{i.e.} to explain why workers work in their employing firms, and conversely, why firms hire some employees and not others. Optimal transport has been used in the economics literature~\cite{lindenlaub2017sorting, galichon2016optimal} to model workers-to-firms matching.

Firms differ in technologies and workers differ in skills. Let $\X \subset \Rp$, $p \geq 1$, denote the set of firms' types (or technologies) and likewise, let $\Y \subset \Rq$, $q \geq 1$, denote the  set of  workers' types in the economy. Given a probability distribution of firm types $\mu \in \PX$ and a probability distribution of worker types $\nu \in \PY$, a coupling $\pi \in \PXY$ of $\mu$ and $\nu$  represents the matching between firms and workers, in the sense that $\pi(A \times B)$, for $A \subset \X, B \subset \Y$ Borel sets, is the proportion of firms whose type is in $A$ that employ a worker whose type is in $B$.
The primal problem is to maximize the total output in the economy:
\begin{align}
    \label{eqn:wot:lindenlaub_ot}
    \OT(\mu,\nu) \defeq \sup_{\pi \in \Pi(\mu,\nu)} \iint_{\XY} F(x,y) \,d\pi(x,y)
\end{align}
where $\Pi(\mu,\nu)$ is the set of all couplings between $\mu$ and $\nu$, \textit{i.e.}
\begin{multline*}
    \Pi(\mu, \nu) = \Big\{ \pi \in \PXY \,,\, \forall A \subset \X, B\subset\Y \text{ Borel},
    \pi(A\times\Y)=\mu(A), \pi(\X\times B)=\nu(B) \Big\},
\end{multline*}
and $F: \XY \to \R$ is the production function, \textit{i.e.} $F(x,y)$ is the output (in $\$$) produced by a worker of type $y \in \Y$ working in a firm of type $x \in \X$.

Problem~\eqref{eqn:wot:lindenlaub_ot} corresponds to the definition of the~\citet{Kantorovich42} problem in optimal transport with cost function $-F$.
It admits the following dual formulation:
\begin{align}
    \label{eqn:wot:kantorovich_dual}
    \OT(\mu,\nu) =
    \inf_{\substack{\chi \in \CX, \phi \in \CY\\\chi \oplus \phi \geq F}}
    \int \chi \,d\mu + \int \phi \,d\nu
\end{align}
where for $\CX$ (\textit{resp.} $\CY$) is the set of real continuous functions over $\X$ (\textit{resp.} over $\Y$), and $\chi \oplus \phi \in \CXY$ is the function $\chi \oplus \phi : (x,y) \mapsto \chi(x) + \phi(y)$. In the labor market  context, $\chi(x)$ and $\phi(y)$ represent respectively the profit
of firms with type $x$ and the wage of workers with type $y$.

\subsection{Weak Optimal Transport}
The Kantorovich problem~\eqref{eqn:wot:lindenlaub_ot} can be rewritten as:
\begin{align}
    \label{eqn:wot:reformulation_kantorovich_wot}
    \sup_{\pi \in \Pi(\mu,\nu)} \int_\X \left[ \int_\Y F(x,y) \,d\pi_x(y) \right] d\mu(x)
\end{align}
where $\left(\pi_x\right)_{x \in \X} \subset \PY$ is the ($\mu$-almost surely unique) probability kernel that allows to disintegrate $\pi$ with respect to $\mu$ as $d\pi(x,y)= d\mu(x) d\pi_x(y)$. In other words, $\pi_x$ is the law of $Y | X=x$ when $(X,Y) \sim \pi$.

\citet{Gozl_Robe_Teta_JFA_2017}~introduce the weak optimal transport  problem as
\begin{align}
    \label{eqn:wot:wot}
    \WOT(\mu,\nu) \defeq \sup_{\pi \in \Pi(\mu,\nu)} \int_\X \F(x, \pi_x) \,d\mu(x)
\end{align}
where $\F: \XPY \to \R$, \textit{i.e.} $\F(x,p)$ now denotes the production (in $\$$) of a firm type $x \in \X$ hiring employees with distribution $p \in \PY$.
The classic  Kantorovich problem is the special case of WOT where $\F(x,p) = \int_\Y F(x,y) \,dp(y)$.

In our economic context, the probability kernel $\pi_x$ represents the distribution of types among the workers hired by firms with types~$x$.
The economic interpretation of the reworded problem~\eqref{eqn:wot:reformulation_kantorovich_wot} is that  the output produced by a  firm of type $x \in \X$ is the sum of the output produced by its employees, $\int F(x,y) \,d\pi_x(y)$. In particular, the production of a firm type $x \in \X$ depends linearly on the distribution of its employees' types, $\pi_x \in \PY$.
\citet{chone2021matching}~relax this restriction and allow firms to aggregate the skills of their employees in more general way. The production of a firm type $x \in \X$  depends non-linearly on the distribution $\pi_x \in \PY$ of its employees'  types.\footnote{WOT is connected to  classic problems in other fields, such as entropic  transport and martingale transport, see Appendix~\ref{app:other}.}

\paragraph{Barycentric WOT problem} \label{par:barycentric_wot} We will say that the WOT problem~\eqref{eqn:wot:wot} is \emph{barycentric} when $\F(x,p)$ only depends on the barycenter of $p$, that is when $\F(x,p) = F\left(x, \int y \,dp(y)\right)$ for some function $F: \XconvY \to \R$, where $\convY$ is the convex hull of $\Y$.\footnote{The particular case where $F(x,y) = \|x-y\|^2$ is called the quadratic barycentric WOT problem.} In the economic context, the barycentric specification is valid if the production of a firm depends 
on the distribution of its employees' types, $p \in \PY$, only through their aggregate skills, $\int y \,dp(y)$.

\subsection{Duality}
\label{subsec:wot:duality}

Just like the Kantorovich problem~\eqref{eqn:wot:lindenlaub_ot} admits the dual formulation~\eqref{eqn:wot:kantorovich_dual}, the WOT problem~\eqref{eqn:wot:wot} also admits a dual formulation under some assumptions on $\F$. For the WOT problem to be convex, and hence hope for strong duality to hold, we require that $p \mapsto \F(x,p)$ is convex for all $x \in \X$. We refer to~\citep[Section 9]{Gozl_Robe_Teta_JFA_2017} for the technical assumptions and details.

Under these assumptions, the WOT problem~\eqref{eqn:wot:wot} admits the following dual formulation~\citep[Theorem 9.5]{Gozl_Robe_Teta_JFA_2017}:
\begin{align}
    \label{eqn:wot:wot_dual}
    \WOT(\mu,\nu) = \inf_{\phi \in \CY} \int_\X R_\F(\phi) \,d\mu + \int_\Y \phi \,d\nu
\end{align}
where $R_\F(\phi)(x) = \sup_{p \in \PY} \F(x,p) - \int \phi \,dp$.

This dual formulation can in turn be interpreted in our economic framework: $\phi(y)$ represents the wage of the worker with type $y \in \Y$. Given a wage function $\phi$, a firm of type $x \in \X$ hires workers according to a probability distribution $p \in \PY$ chosen to maximize profit defined as output $\F(x,p)$ minus wage bill $\int \phi(y) \,dp(y)$. $R_\F(\phi)(x)$ is therefore the maximum profit the firm type $x \in \X$ can attain given the wage function $\phi$, so that $\int_\X R_\F(\phi) \,d\mu$ is the total profit in the economy. The salaries are then chosen so as to minimize the sum of the profits and of the salaries.

When the cost function $\F$ is barycentric, \textit{i.e.} when $\F(x,p) = F\left(x, \int y \,dp(y)\right)$ for some $F: \XconvY \to \R$
\citep[Proof of Theorem 2.11]{Gozl_Robe_Teta_JFA_2017} prove another dual formulation:
\begin{align}
    \label{eqn:wot:barycentric_wot_dual}
    \WOT(\mu,\nu) =
    \inf_{\substack{\psi \in \CconvY\\\text{convex, Lipschitz}}}
    \int_\X Q_F(\psi) \,d\mu + \int_\Y \psi \,d\nu
\end{align}
where $Q_F(\psi)(x) = \sup_{y \in \convY} F(x,y) - \psi(y) $.
\citep[Proof of Theorem 2.11]{Gozl_Robe_Teta_JFA_2017} gives a way to construct a minimizer $\opt{\psi}$ of the dual~\eqref{eqn:wot:barycentric_wot_dual} from a minimizer $\opt{\phi}$ of the more general dual problem~\eqref{eqn:wot:wot_dual}, by simply taking for $\opt{\psi}$ the largest convex function that is smaller than $\opt{\phi}$, \textit{i.e.}:
\begin{align}
    \label{eqn:wot:phi_to_psi}
    \opt{\psi}: z \mapsto \inf_{\substack{p \in \PY\\\int y \,dp(y) = z}} \int_\Y \opt{\phi} \,dp.
\end{align}

The convexity of dual minimizers in the barycentric case is easy to interpret in our economic setting.
Here the output produced by a firm depends only on the aggregate skill of its employees. If the wage function is $\opt{\phi}$, $\opt{\psi}(z)$ given by~\eqref{eqn:wot:phi_to_psi} represents the lowest wage bill that a firm must spend to achieve the aggregate skill $z=\int y\, dp(y)$.
The convexity of the wage thus directly results from the firms' ability to aggregate the skills of their employees. 


\section{Weak Optimal Transport with Unnormalized Kernel (WOTUK)} \label{sec:wotuk}

\subsection{From WOT to WOTUK}
\label{sec:WOT:WOTUK}

\citet{chogozkra2021}~relax the assumption that $\pi_x$ in~\eqref{eqn:wot:reformulation_kantorovich_wot} and~\eqref{eqn:wot:wot} is a
\emph{probability} measure. They allow $\pi_x$ to be a positive measure. Denoting by $\MY$ the set of positive measures over $\Y$, they introduce the weak optimal transport problem with unnormalized kernel (WOTUK) as
\begin{align}
    \label{eqn:wotuk:wotuk}
    \WOTUK(\mu,\nu) \defeq \sup_{\substack{q \in \MY^\X\\\int q_x \,d\mu(x) = \nu}} \int_\X \F(x, q_x) \,d\mu(x)
\end{align}
where $\F: \XMY \to \R$. The constraint $\int q_x \,d\mu(x) = \nu$ expresses that the unnormalized kernel $q$ transports $\mu$ onto $\nu$.
\citet{chogozkra2021}~connect the WOTUK problem~\eqref{eqn:wotuk:wotuk} to a WOT problem as follows.
Letting
\[
\Pi(\ll \mu, \nu) \defeq \{ \pi \in \Pi(\eta, \nu) \,,\, \eta \in \PX, \eta \ll \mu \},
\]
denote the set of probability measure over $\X$ that are absolutely continuous with respect to $\mu$, they show that
\begin{align}
    \WOTUK(\mu,\nu) 
    &= \sup_{\substack{\eta \in \PX\\\eta \ll \mu}}
    \sup_{\pi \in \Pi(\eta, \nu)}
    \int \F\left(x, \frac{d\eta}{d\mu}(x) \pi_x \right) \,d\mu(x) 
    \label{eqn:wotuk:wotuk_with_eta}\\
    &= \sup_{\pi \in \Pi(\ll \mu, \nu)} \int \F\left(x, \frac{d\pi_1}{d\mu}(x) \pi_x \right) \,d\mu(x)
\end{align}
where $\pi_x \in \PY$ is the unique disintegration of $\pi$ with respect to $\eta$, \textit{i.e.} such that $d\pi(x,y) = d\eta(x) d\pi_x(y)$, and $\pi_1$ is the first marginal of $\pi$.
At given $\eta$, we thus get back the WOT problem studied in Section~\ref{sec:wot}. Instead of constraining the first marginal of $\pi$ to be $\mu$, the WOTUK problem only imposes that the first marginal is absolutely continuous with respect to $\mu$.
\citet{chogozkra2021}~show that the density of $\eta$ with respect to $\mu$ is nothing else than the mass of $q_x$, \textit{i.e.}, $\frac{d\eta}{d\mu}(x)=\frac{d\pi_1}{d\mu}(x)=q_x(\Y)$.

In the economic setting of~\cite{chone2021matching}, $q_x(\Y)$ represents the number of employees (\textit{i.e.}, the size) of firms with type~$x$. Allowing $q_x$ to be an unnormalized positive measure instead of a probability measure avoids having to assume that all firms have the same size. In contrast to earlier literature, firms' sizes are unknowns to be determined rather than given parameters.


\paragraph{Conical WOTUK problem}
The conical WOTUK problem corresponds to the case where
\[
    \F(x,p) = F\left(x, \int_\Y y \,dp(y)\right)
\]
for some $F: \XconeY \to \R$, where the conical hull of $\Y$ is given by
\[
    \coneY \defeq
    \Bigg\{ \sum_{i=1}^n \lambda_i y_i \,,\, \lambda_1, \ldots, \lambda_n \in \R_+,
    y_1, \ldots, y_n \in \Y, n \geq 1 \Bigg\}.
\]
In \cite{chone2021matching}, a firm's output depends on the \emph{conical} combination of its employees' types, $\int y \,dq_x(y)$. The combination is said to be ``conical'' because the mass of $q_x$ is not necessarily equal to one. In other words, the aggregate skill of the workers  hired by a firm is not their average skills as in the WOT setting, but their average skills \emph{scaled by the positive factor} $q_x(\Y)$ that represents the number of employees.


\subsection{Duality}
The WOTUK problem~\eqref{eqn:wotuk:wotuk} admits dual formulations that are similar to the dual WOT formulations~\eqref{eqn:wot:wot_dual} and \eqref{eqn:wot:barycentric_wot_dual}. The main difference with the results of subsection~\ref{subsec:wot:duality} lies in the fact that $\PY$ is replaced by $\MY$ and that $\convY$ is replaced by $\coneY$. 

Under some technical assumptions on $\F$ detailed in~\cite{chogozkra2021}, the theorem 3.2 in the same reference proves that the WOTUK problem~\eqref{eqn:wotuk:wotuk} admits the following dual formulation:
\begin{align}
    \label{eqn:wotuk:wotuk_dual}
    \WOTUK(\mu,\nu) =
    \inf_{\phi \in \CbY} \int_\X K_\F(\phi) \,d\mu + \int_\Y \phi \,d\nu
\end{align}
where $K_\F(\phi)(x) = \sup_{m \in \MY} \F(x,m) - \int \phi \,dm$.

Similarly, \citep[Theorem 5.1]{chogozkra2021}~proves that the conical WOTUK problem admits the dual formulation:
\begin{align}
    \label{eqn:wotuk:barycentric_wotuk_dual}
    \WOTUK(\mu,\nu) =
    \inf_{\substack{\psi \in \CconeY \text{ convex, }\\\text{positively homogeneous}}}
    \int_\X J_F(\psi) \,d\mu + \int_\Y \psi \,d\nu
\end{align}
where $J_F(\psi)(x) = \sup_{y \in \coneY} F(x,y) - \psi(y) $.
They show that a minimizer $\opt{\psi}$ of the dual problem~\eqref{eqn:wotuk:barycentric_wotuk_dual} from a minimizer $\opt{\phi}$ of the more general dual problem~\eqref{eqn:wotuk:wotuk_dual} by  taking for $\opt{\psi}$ the largest convex and positively homogeneous function that is smaller than $\opt{\phi}$, \textit{i.e.}:
\begin{align}
    \label{eqn:wotuk:phi_to_psi}
    \opt{\psi}: z \mapsto \inf_{\substack{m \in \MY\\\int y \,dm(y) = z}} \int_\Y \opt{\phi} \,dm.
\end{align}

In the economic setting of \cite{chone2021matching}, a dual optimizer $\phi$ is  a wage function: $\phi(y)$ represents the wage paid to a worker of type $y \in \Y$.
As $R_\F(\phi)(x)$ and $Q_F(\psi)(x)$ in subsection~\ref{subsec:wot:duality}, $K_\F(\phi)(x)$ and $J_F(\psi)(x)$ are two forms for the profit function, \textit{i.e.}, for the maximal profit that firms of each type $x\in\X$ achieve under the wage functions $\phi$ or $\psi$.

\section{Algorithms} \label{sec:algo}

In this section, we only consider the case of discrete measures $\mu \in \PX$ and $\nu \in \PY$. We will write $\mu = \sum_{i=1}^n a_i \delta_{x_i}$ where $n \geq 1$ is the number of firm types, $x_1, \ldots, x_n \in \X$ are the firm types and $a \in \R^n$ represents the proportion of the firm types in the economy ($a > 0$ and $\sum_{i=1}^n a_i = 1$). Likewise, we will write $\nu = \sum_{j=1}^m b_j \delta_{y_j}$ where $m \geq 1$ is the number of worker types, $y_1, \ldots, y_m \in \Y$ are the worker types and $b$ represents the proportions of the worker types in the population ($b > 0$ and $\sum_{j=1}^m b_j = 1$). Since we consider here  maximization problems, we will require that the cost function $\F$ is concave and differentiable with respect to its second argument.

\subsection{The primal problems}

A matching $\pi \in \Pi(\mu,\nu)$ is now represented by a matrix $P \in \Rnm$ such that $P_{ij}$ represents the proportion of the firm type $x_i$ which is matched with the worker type $y_j$. For the WOT problem~\eqref{eqn:wot:wot}, the marginal constraints on $P$ translate into:
\[
    \Pi(\mu,\nu) = \{ P \in \Rnm_+ \,,\, P \ones = a, P^\top \ones = b \}.
\]
For the WOTUK problem~\eqref{eqn:wotuk:wotuk_with_eta}, the set of constraints is simply
\[
\Pi(\ll \mu, \nu) = \{ P \in \Rnm_+ \,,\, P^\top \ones = b\}
\]
because as explained in Section~\ref{sec:WOT:WOTUK} the first marginal $P \ones = \eta$ is unconstrained and $\eta$ is an unknown variable to be determined.
The difference between the WOT problem~\eqref{eqn:wot:wot} and the WOTUK problem~\eqref{eqn:wotuk:wotuk_with_eta} lies in the constraint set only. The WOTUK problem corresponds to the WOT problem where the first marginal constraint has been removed, which establishes an interesting link with the unbalanced optimal transport theory~\cite{2015-chizat-unbalanced}.

Let us now write the objective for the WOT and WOTUK problems in the discrete setting. The disintegration $\pi_{x_i}$ representing the workers hired by the firm of type $x_i$ writes $\frac{1}{a_i} \sum_{j=1}^m P_{ij} \delta_{y_j}$. For simplicity, we make the following change of notations: we define $\algoF: \X\times\R^m$ by
\[
    \algoF : (x,p) \mapsto \F\left(x, \sum_{j=1}^m p_j \delta_{y_j}\right).
\]
Note that $\algoF$ depends on $y_1, \ldots, y_m$ and that $\frac{\partial \algoF}{\partial p_j}(x, p)
= \left\langle \delta_{y_j}, \nabla_2 \F\left(x, \sum_{k=1}^m p_k \delta_{y_k} \right) \right\rangle $ where $\langle \cdot, \cdot \rangle: \MX \times \MX^*$ is the duality bracket. We will use the notation $P_{i:} = (P_{ij})_{1 \leq j \leq m}$ for $1 \leq i \leq n$.

With these notations, the objective of the WOT problem~\eqref{eqn:wot:wot} and of the WOTUK problem~\eqref{eqn:wotuk:wotuk_with_eta} writes:
\begin{align*}
    \!\!\!
    f(P) \defeq
    \sum_{i=1}^n a_i \F\left(x_i, \frac{1}{a_i} \sum_{j=1}^m P_{ij} \delta_{y_j}\right)
    = 
    \sum_{i=1}^n a_i \algoF\left(x_i, \frac{P_{i:}}{a_i} \right).
\end{align*}

Since the constraints sets $\Pi(\mu, \nu)$ and $\Pi(\ll \mu, \nu)$ are convex and $\F$ and $\algoF$ are convex in their second argument, both the discrete WOT and WOTUK problems are convex (although not strictly) optimization problems. In order to solve them, we propose to apply a mirror ascent on $P$ (using the Kullback-Leibler divergence). The gradient of the total output writes:
\[
    \frac{\partial f}{\partial P_{ij}}(P) = \left[ \nabla_2 \algoF\left(x_i, \frac{P_{i:}}{a_i} \right) \right]_j.
\]
After each gradient step, the resulting matching $P$ should be projected (for the KL divergence) onto $\Pi(\mu,\nu)$ (for the WOT problem) or $\Pi(\ll \mu,\nu)$ (for the WOTUK problem). For the WOT problem, we have to solve $\min_{Q \in \Pi(\mu,\nu)} \KL(Q | P)$. This problem is equivalent to the entropic OT problem~\eqref{eqn:wot:entropic_ot} with cost function $-\log P$ and regularization strength $\epsilon=1$, and can therefore be efficiently solved using the Sinkhorn algorithm~\cite{CuturiSinkhorn}. For the WOTUK problem, we have to solve $\min_{Q \in \Pi(\ll \mu,\nu)} \KL(Q | P)$ which admits the following closed-form solution (see a proof in Appendix~\ref{appendix:proof:projection_dual_wotuk}): $\opt{Q} = P \odot b / P^\top\ones$ where $\odot$ and $/$ are the elementwise multiplication and division.

\begin{wrapfigure}{R}{0.45\textwidth}
    \vskip-0.75cm
    \begin{minipage}{0.45\textwidth}
        \begin{algorithm}[H]
        	\caption{Mirror Ascent Algorithm for WOT and WOTUK (primal)}
        	\label{algo:primal:wot_wotuk}
        	\begin{algorithmic}
        	    \STATE{Input} Stepsize $\gamma > 0$, tolerance $\epsilon$
        		\STATE{Initialize} $P = a b^\top$
           		\WHILE{$\Ugap(P) > \epsilon f(P)$}
        		\STATE $P \leftarrow P \exp\left(\gamma \, \nabla f(P)\right)$
        		\STATE \textit{\underline{For WOT:}}
        		\STATE \quad $P \leftarrow \text{Sinkhorn}(a,b,\text{kernel}=P)$
        		\STATE \textit{\underline{For WOTUK:}}
        		\STATE \quad $P \leftarrow P \odot b / P^\top\ones$
          		\ENDWHILE
          		\STATE{Return} $P$
        	\end{algorithmic}
        \end{algorithm}
    \end{minipage}
    \vskip-0.7cm
\end{wrapfigure}

\paragraph{Numerical guarantee} Along the mirror ascent iterations over $P$, we can monitor the convergence by looking at the gap
\[
    \gap(P) \defeq \sup_{Q \in K} f(Q) - f(P)
\]
where $K = \Pi(\mu,\nu)$ for the WOT problem and $K = \Pi(\ll\mu,\nu)$ for the WOTUK problem. By definition, $\gap(P) \geq 0$ and by the concavity of $f$,
\[
    \gap(P) \leq \sup_{Q \in K} \langle \nabla f(P), Q-P \rangle \defeq \Ugap(P).
\]
When $K = \Pi(\mu,\nu)$, the upper bound $\Ugap(P)$ on $\gap(P)$ corresponds to an optimal transport problem (with cost matrix $-\nabla f(P)$) and can either be computed exactly or be itself upper bounded using the solution $\opt{Q}$ of an entropic OT problem (efficiently solved using the Sinkhorn algorithm). When $K = \Pi(\ll\mu,\nu)$, the upper bound $\Ugap(P)$ on $\gap(P)$ admits the following closed form solution: $\opt{Q}_{ij} = b_j$ if $i = \argmax_{1 \leq k \leq n} [\nabla f(P)]_{kj}$ and $\opt{Q}_{ij} = 0$ otherwise.

The algorithm stops when $\Ugap(P) \leq \epsilon f(P)$ for a tolerance $\epsilon>0$. We summarize the mirror ascent method for the WOT and WOTUK problems in Algorithm~\ref{algo:primal:wot_wotuk}.

\newpage
\subsection{The dual problems}

In the discrete setting, the general dual for WOT~\eqref{eqn:wot:wot_dual} writes:
\begin{align}
    \label{eqn:algo:wot_dual}
    \min_{\phi \in \Rm_+} 
    \langle b, \phi \rangle + 
    \max_{\substack{P \in \Rnm_+\\P \ones = a}}
    \sum_{i=1}^n a_i \algoF\left(x_i, \frac{P_{i:}}{a_i} \right) - \ones^\top P \phi
\end{align}
and likewise for the general dual for the WOTUK problem~\eqref{eqn:wotuk:wotuk_dual}:
\begin{align}
    \label{eqn:algo:wotuk_dual}
    \min_{\phi \in \Rm_+} 
    \langle b, \phi \rangle + 
    \max_{P \in \Rnm_+}
    \sum_{i=1}^n a_i \algoF\left(x_i, \frac{P_{i:}}{a_i} \right) - \ones^\top P \phi.
\end{align}

Let us define
\[
    h : (\phi, P) \mapsto \sum_{i=1}^n a_i \algoF\left(x_i, \frac{P_{i:}}{a_i} \right) - \ones^\top P \phi.
\]

\begin{wrapfigure}{R}{0.50\textwidth}
    \vskip-0.7cm
    \begin{minipage}{0.50\textwidth}
        \begin{algorithm}[H]
        	\caption{Mirror Descent Algorithm for WOT and WOTUK (dual)}
        	\label{algo:dual:wot_wotuk}
        	\begin{algorithmic}
        	    \STATE{Input} $\gamma_1 > 0$, $\gamma_2 > 0$, $K_1, K_2 \in \N$
        		\STATE{Initialize} $\phi \in \Rm_+$ and $P = a b^\top$
           		\FOR{$k_1=0$ {\bfseries to} $K_1$}
           		    \FOR{$k_2=0$ {\bfseries to} $K_2$}
           		        \STATE $P \leftarrow P \exp\left(\gamma_1 \, \nabla_P h(\phi,P)\right)$
        		        \STATE \textit{\underline{For WOT:}}
           		        \STATE \quad $P \leftarrow \diag\left( a / P\ones \right) P$
          		    \ENDFOR
          		    \STATE $\phi \leftarrow \phi \exp\left( - \gamma_2 \left[b - P^\top \ones\right] \right)$
          		\ENDFOR
          		\STATE Use a linear programming solver to compute $\psi(z)$ for $z \in \Y$:
          		\STATE \textit{\underline{For WOT:}}
          		\[
          		    \psi(z) = \max_{\substack{\lambda \in \mathbb{R}^q, \mu \in \R \\ \forall j, \langle \lambda, y_j \rangle + \mu \leq \phi_j}} \langle \lambda, z \rangle + \mu.
          		\]
          		\STATE \textit{\underline{For WOTUK:}}
          		\[
          		    \psi(z) = \max_{\substack{\lambda \in \mathbb{R}^q \\ \forall j, \langle \lambda, y_j \rangle \leq \phi_j}} \langle \lambda, z \rangle.
          		\]
          		\STATE{Return} Dual variables $\phi_j, \psi(y_j)$ for $1 \leq j \leq m$.
        	\end{algorithmic}
        \end{algorithm}
    \end{minipage}
    \vskip-0.5cm
\end{wrapfigure}
To solve problems~\eqref{eqn:algo:wot_dual} and \eqref{eqn:algo:wotuk_dual}, we propose to run a mirror descent on $\phi$ (with the Kullback-Leibler divergence). The objective is itself a maximization problem (over $P$). The envelope theorem yields the gradient of the objective, provided the optimal $P$ is given. At each gradient step on $\phi$, we therefore propose to run a mirror ascent on $P$ at fixed $\phi$. For the WOTUK problem~\eqref{eqn:algo:wotuk_dual}, no projection are needed, while for the WOT problem~\eqref{eqn:algo:wot_dual}, we need to project $P$ onto $\{P \in \Rnm_+, P\ones=a\}$ during the ascents. The proof in Appendix~\ref{appendix:proof:projection_dual_wotuk} directly adapts to this case, and the projection amounts to reweighting the rows of $P$.

We can construct dual minimizers for the barycentric WOT~\eqref{eqn:wot:barycentric_wot_dual} and conical WOTUK~\eqref{eqn:wotuk:barycentric_wotuk_dual} from a solution $\opt\phi$ of~\eqref{eqn:algo:wot_dual} and \eqref{eqn:algo:wotuk_dual} respectively, using the results given in~\eqref{eqn:wot:phi_to_psi} and \eqref{eqn:wotuk:phi_to_psi} respectively. In the discrete setting, these linear programs respectively write:
\[
    \opt{\psi}: z \mapsto  \min_{\substack{p \in \mathbb{R}_+^m \\ \sum_{j=1}^m p_j = 1 \\ \sum_{j=1}^m p_j y_j = z}} \langle p, \opt\phi \rangle
\]
and for WOTUK
\[
    \opt{\psi}: z \mapsto  \min_{\substack{p \in \mathbb{R}_+^m \\ \sum_{j=1}^m p_j y_j = z}} \langle p, \opt\phi \rangle.
\]
Since we may be interested in differentiating those functions $\opt\psi$, we rather compute the dual problems of the above linear programs (see a proof in the Appendix~\ref{appendix:proof:phi_to_psi}):
\[
    \opt\psi: z \mapsto \max_{\substack{\lambda \in \mathbb{R}^q, \mu \in \R \\ \forall j, \langle \lambda, y_j \rangle + \mu \leq \opt{\phi}_j}} \langle \lambda, z \rangle + \mu
    \quad \text{and} \quad
    \opt\psi: z \mapsto \max_{\substack{\lambda \in \mathbb{R}^q \\ \forall j, \langle \lambda, y_j \rangle \leq \opt{\phi}_j}} \langle \lambda, z \rangle.
\]
We summarize the mirror ascent method for the WOT and WOTUK problems in Algorithm~\ref{algo:dual:wot_wotuk}.
\section{Experiments and Applications in Economics} \label{sec:expes}

All the experiments have been run on a Google colab notebook using JAX. We used the following packages: OTT~\cite{OTT} for entropic optimal transport, POT~\cite{flamary2021pot} for exact optimal transport, GLPK~\cite{makhorin2008glpk} through CVXPY~\cite{agrawal2018rewriting} for linear programming.

\begin{figure}[h]
    \vskip -1.5cm
    \centering
    \includegraphics[width=0.7\textwidth]{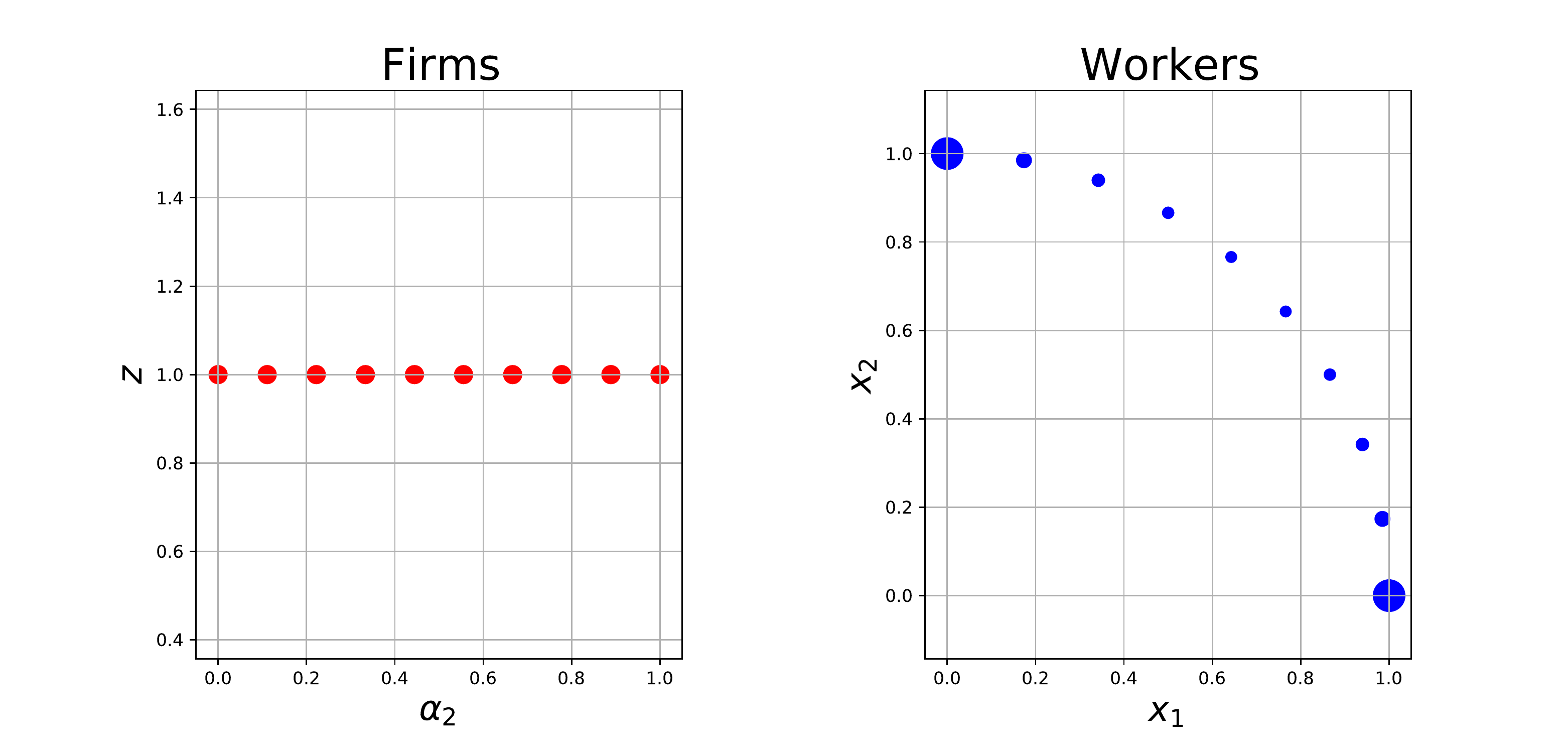}
    \caption{An example of discrete probability distributions of firm types $\mu$ (left) and worker types $\nu$ (right). Each dot represents a Dirac mass, and its size represents its weight. Here, all firm types have the same weight while there are more specialist workers in the economy than generalists.}
    \label{fig:expes:marginals}
    \vskip -0.5cm
\end{figure}

From now on, we focus on the setting of~\cite{chone2021matching} and will therefore consider the distributions of firm types and worker types and the cost function they use in their model. We focus on the case where there are two skills in the economy. The set of firm types is $\X = \{(z, \alpha_1, \alpha_2) \in \R_+^3, \alpha_1 + \alpha_2 = 1\}$ where $z$ represents the productivity of the firm type and $\alpha_i$, $i \in \{1,2\}$ its demand in skill $i$. The set of worker types is $\Y = \{(x_1, x_2) \in \R_+^2\}$, where $x_i$, $i \in \{1,2\}$, represents the proficiency of the worker type in skill $i$. We will denote the skill profile of a worker type by $\theta = \arctan\left(x_2/x_1\right) \in [0, \frac{\pi}{2}]$, where $\theta=0$ represents an expert worker in skill $1$, $\theta=\frac{\pi}{2}$ an expert worker in skill $2$ and  $\theta=\frac{\pi}{4}$ a generalist worker. 
The skill profile represents the worker's comparative advantage in skill 2 over skill 1.
We plot in Figure~\ref{fig:expes:marginals} an example of such distributions.

As for the production function, we consider a Constant Elasticity of Substitution (CES) function:
\[
    F\left((z, \alpha_1, \alpha_2), (x_1, x_2)\right) = \frac{z}{\zeta} \left( \alpha_1 x_1^\sigma + \alpha_2 x_2^\sigma \right)^{\zeta/\sigma}
\]
with parameters $\zeta=\sigma=\frac{1}{2}$.

\subsection{Comparison between OT, EOT, WOT and WOTUK}

\begin{figure}[!h]
     \centering
     \begin{subfigure}[b]{0.23\textwidth}
         \centering
         \includegraphics[width=\textwidth, trim = {0 1.6cm 0 1.5cm}, clip]{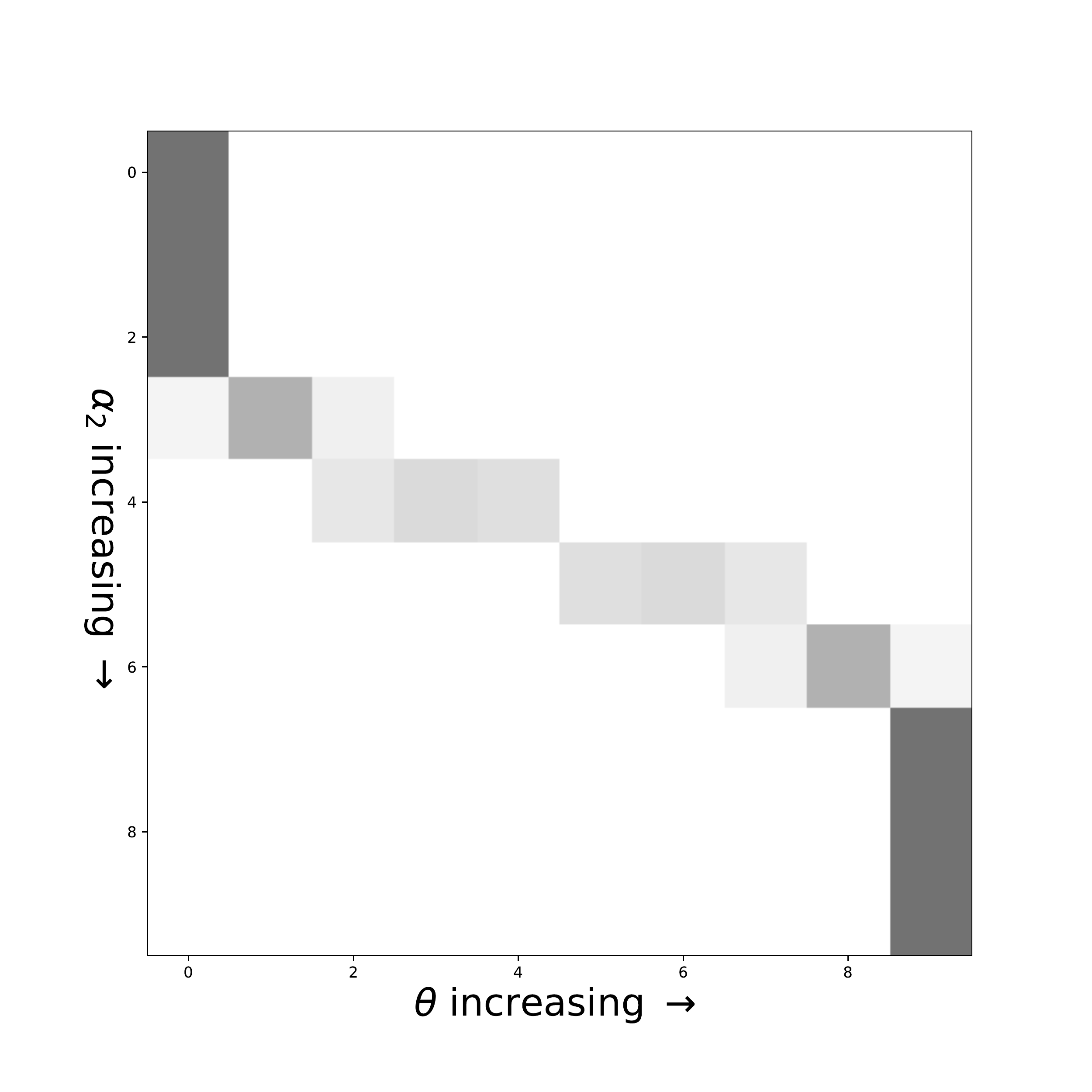}
         \caption{OT}
         \label{fig:expes:OTplan}
     \end{subfigure}
     \begin{subfigure}[b]{0.23\textwidth}
         \centering
         \includegraphics[width=\textwidth, trim = {0 1.6cm 0 1.5cm}, clip]{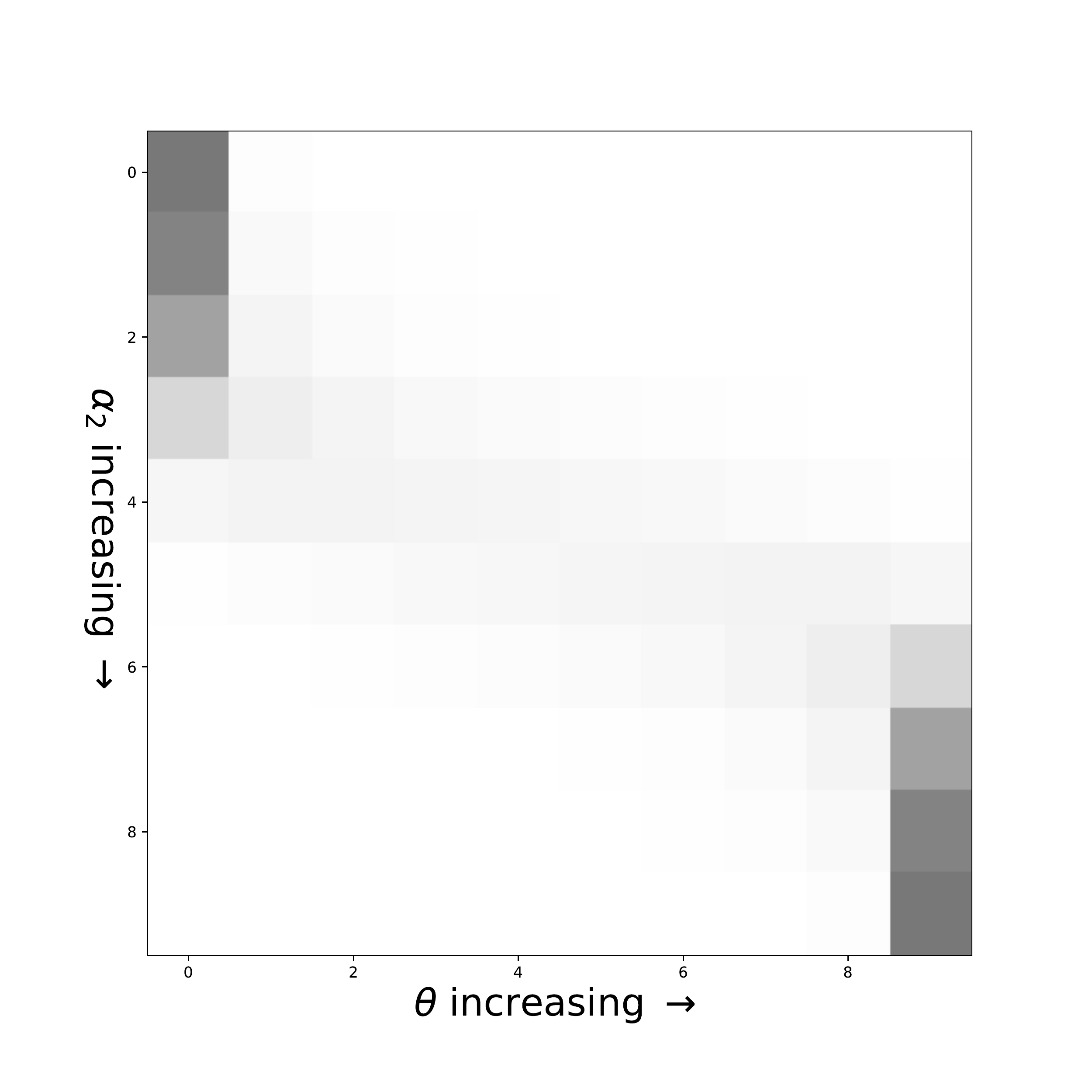}
         \caption{EOT ($\epsilon=0.1$)}
         \label{fig:expes:Sinkhornplan}
     \end{subfigure}
     \begin{subfigure}[b]{0.23\textwidth}
         \centering
         \includegraphics[width=\textwidth, trim = {0 1.6cm 0 1.5cm}, clip]{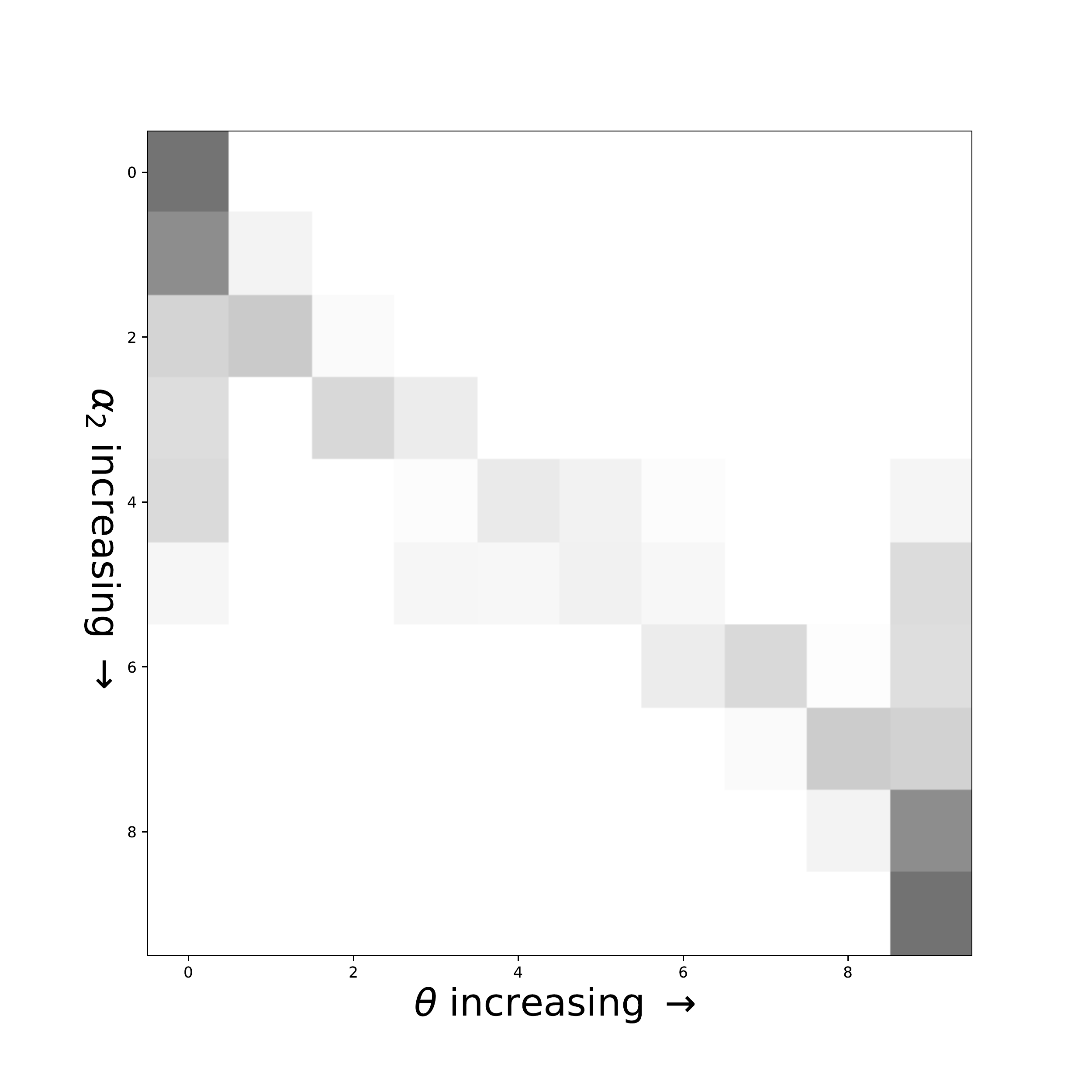}
         \caption{WOT}
         \label{fig:expes:WOTplan}
     \end{subfigure}
     \begin{subfigure}[b]{0.23\textwidth}
         \centering
         \includegraphics[width=\textwidth, trim = {0 1.6cm 0 1.5cm}, clip]{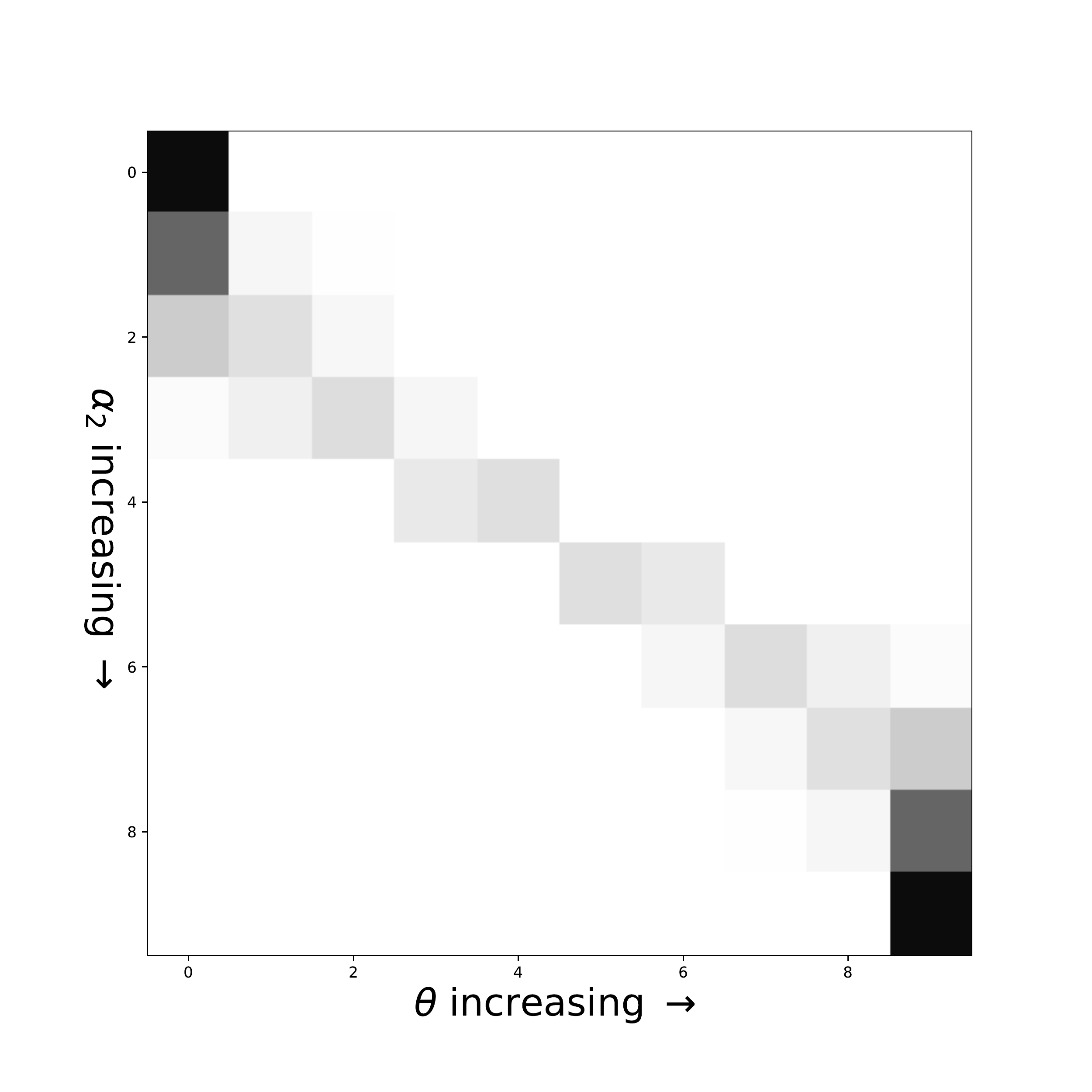}
         \caption{WOTUK}
         \label{fig:expes:WOTUKplan}
     \end{subfigure}
        \caption{Optimal transport plans for different variants of OT, between the marginals depicted in Figure~\ref{fig:expes:marginals}. Each row represents a firm type, each column represents a worker type, and each cell represents the matching strength between the corresponding firm and worker types. The darker the cell, the stronger the matching. The color scale is the same in the four pictures.}
        \label{fig:expes:transport_plans}
\end{figure}

\begin{wrapfigure}{R}{0.45\textwidth}
    \vskip -0.8cm
    \centering
    \includegraphics[width=0.45\textwidth]{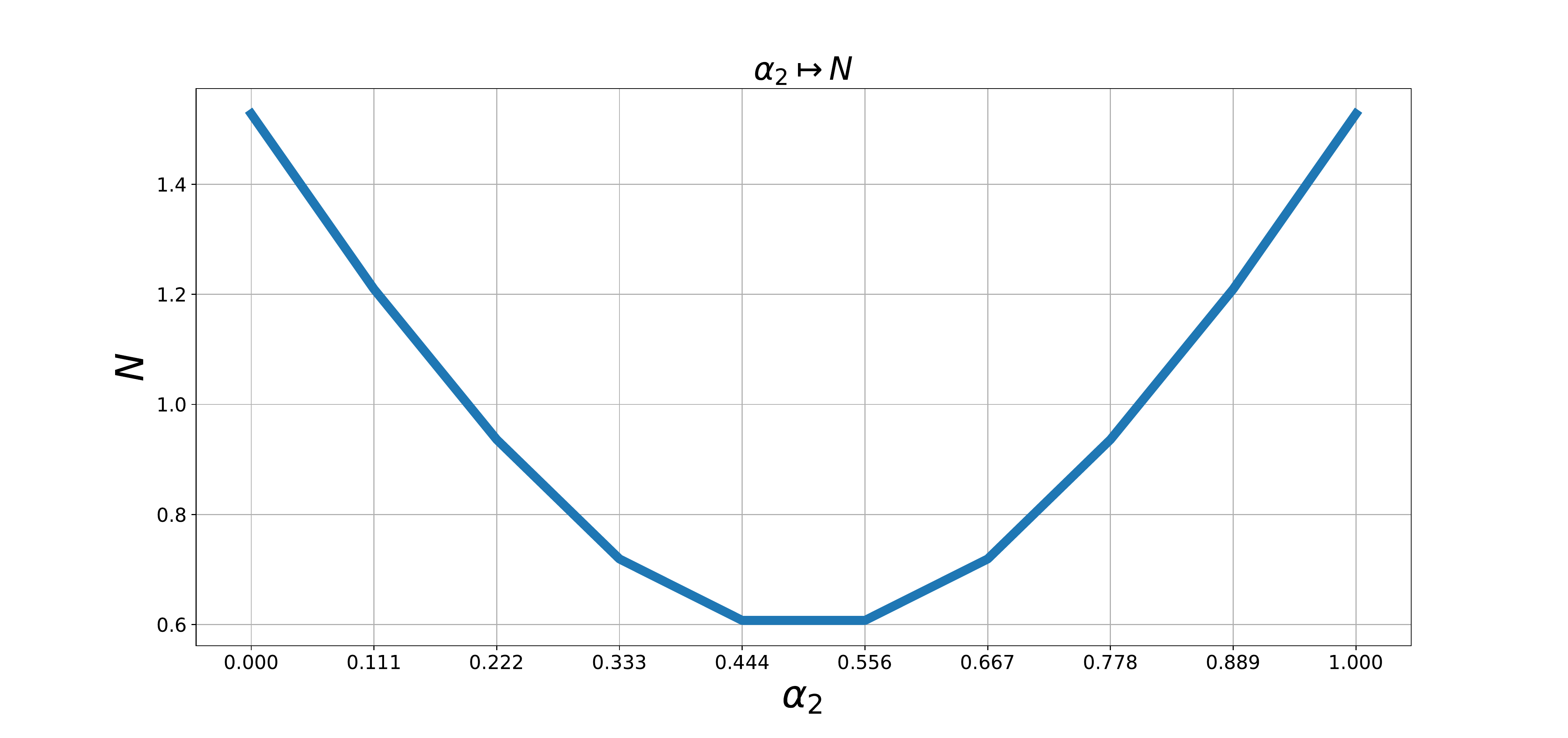}
    \caption{\looseness=-1 The firm types' size $N(z,\alpha_1,\alpha_2) = \frac{d\eta}{d\mu}(z,\alpha_1,\alpha_2)$ in function of $\alpha_2$ in the WOTUK model. See equation~\eqref{eqn:wotuk:wotuk_with_eta} for the definition of $\eta$. Specialist firms get bigger while generalist firms get smaller.}
    \label{fig:expes:alpha_to_N}
    \vskip -0.8cm
\end{wrapfigure}

In Figure~\ref{fig:expes:transport_plans}, we showcase the empirical differences between four different optimal transport problems, when computed on the economic setting described above: the OT problem~\eqref{eqn:wot:lindenlaub_ot}, the EOT problem~\eqref{eqn:wot:entropic_ot}, the WOT problem~\eqref{eqn:wot:wot} and the WOTUK problem~\eqref{eqn:wotuk:wotuk}.

While all four methods agree that firms with a higher demand in skill $2$ (\textit{i.e.} with a higher value of $\alpha_2$) employ workers with better proficiency in skill $2$ (\textit{i.e.} with higher $\theta$), we can spot some differences.

The OT and WOTUK plans (Figures~\ref{fig:expes:OTplan}~and~\ref{fig:expes:WOTUKplan}) are sparse, \textit{i.e.} firms do not tend to employ workers of different types, while the EOT and WOT plans (Figure~\ref{fig:expes:Sinkhornplan}~and~\ref{fig:expes:WOTplan}) show that in these models, generalist firms (\textit{i.e.} firms such that $\alpha_1 \approx \alpha_2$) mix different types of workers. Since the EOT plan is a blurry version of the OT plan, firms in this model mix workers of different but similar types. On the other hand, generalist firms in the WOT model employ both specialist workers of both skills, and generalist workers.

Since there are ten firm types with uniform proportion in the economy, the maximal value in the transport matrix is $1/10$ in the OT, EOT and WOT models. Since the WOTUK model is unnormalized, its optimal plan may (and does) display higher values. We depict in Figure~\ref{fig:expes:alpha_to_N} the size of firm types in function of their $\alpha_2$. Specialist firms get bigger and generalist firms get smaller, which was to be expected since there are more specialist workers and less generalist workers on the job market.

\subsection{Simulation of the economic model}


\begin{wrapfigure}{R}{0.45\textwidth}
    \vskip-0.4cm
     \centering
     \begin{subfigure}[b]{0.450\textwidth}
         \centering
         \includegraphics[width=\textwidth, trim = {1.6cm 0 1.6cm 0}, clip]{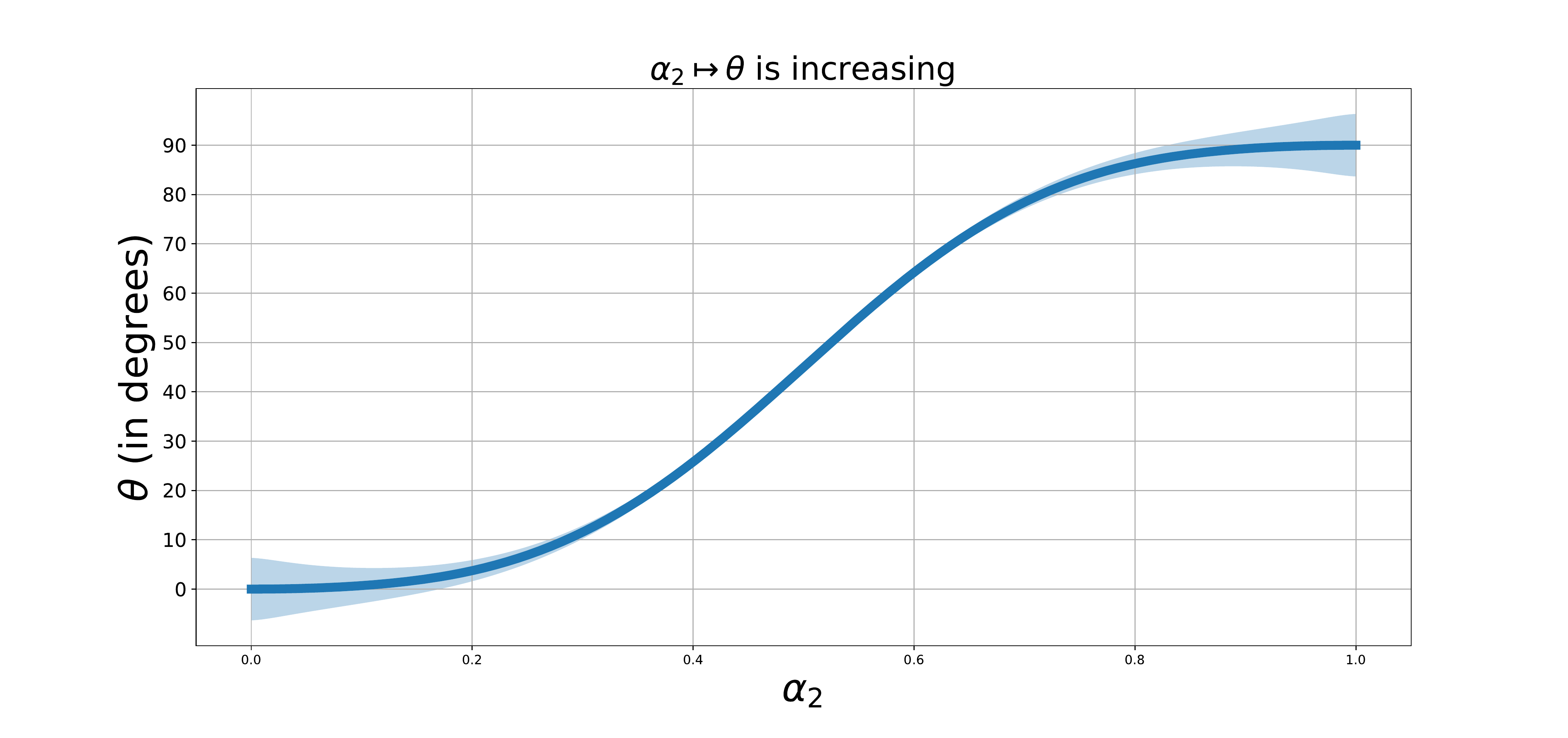}
         \caption{Scenario A}
         \label{fig:expes:eco:alphatotheta_specialists}
     \end{subfigure}
     \begin{subfigure}[b]{0.450\textwidth}
         \centering
         \includegraphics[width=\textwidth, trim = {1.6cm 0 1.6cm 0}, clip]{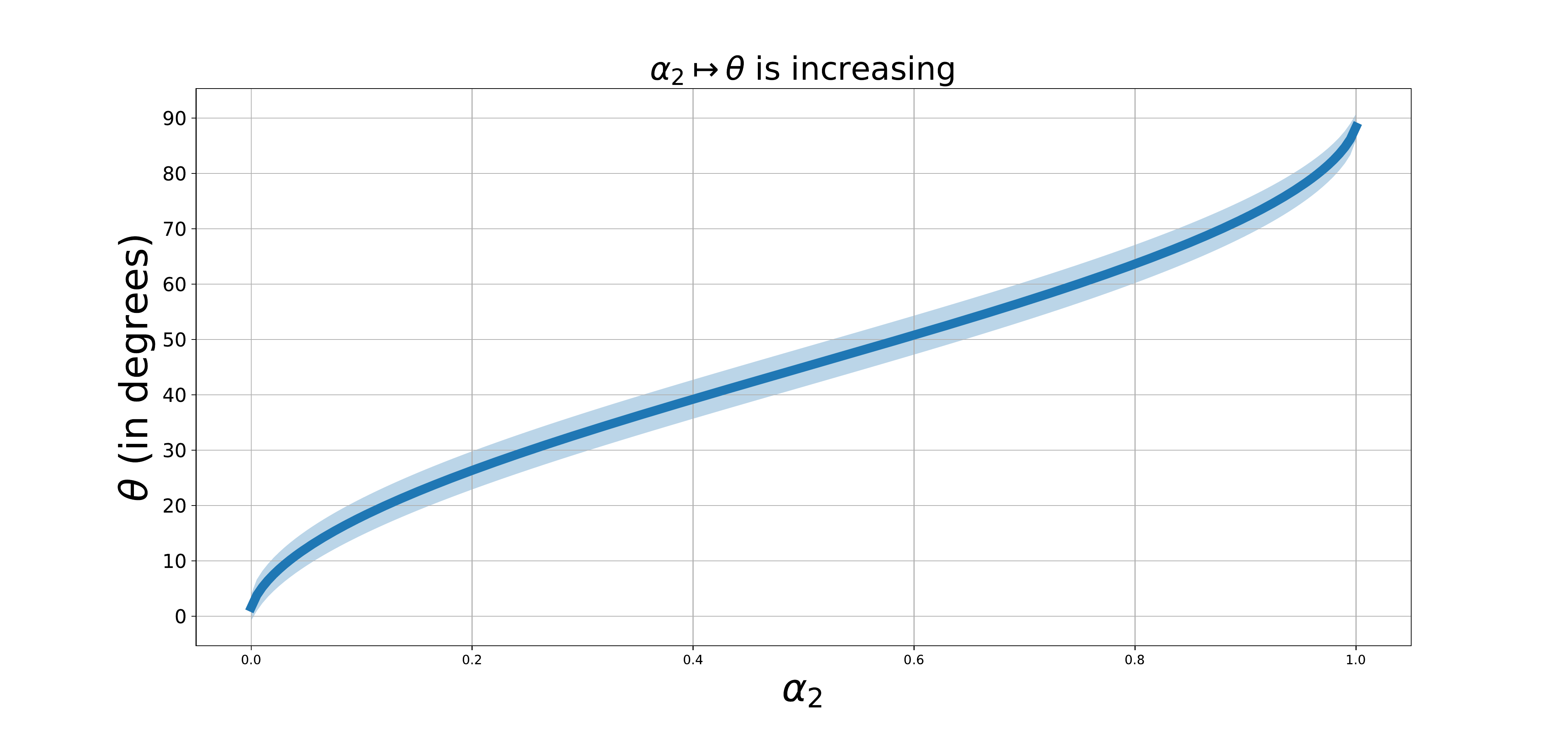}
         \caption{Scenario B}
         \label{fig:expes:eco:alphatotheta_generalists}
     \end{subfigure}
        \caption{The ``mapping'' $\alpha_2 \mapsto \theta$.}
        \label{fig:expes:eco:alpha_to_theta}
\end{wrapfigure}

We now consider the same 2D model, but with 200 firms and workers. Like before, the firms are distributed uniformly along $z$, while we consider the two cases: many specialist workers (scenario A) and many generalist workers (scenario B).

\begin{figure}[h!]
    \vskip-0.3cm
     \centering
     \begin{subfigure}[b]{0.30\textwidth}
         \centering
         \includegraphics[width=\textwidth, trim = {0 1.6cm 0 1.5cm}, clip]{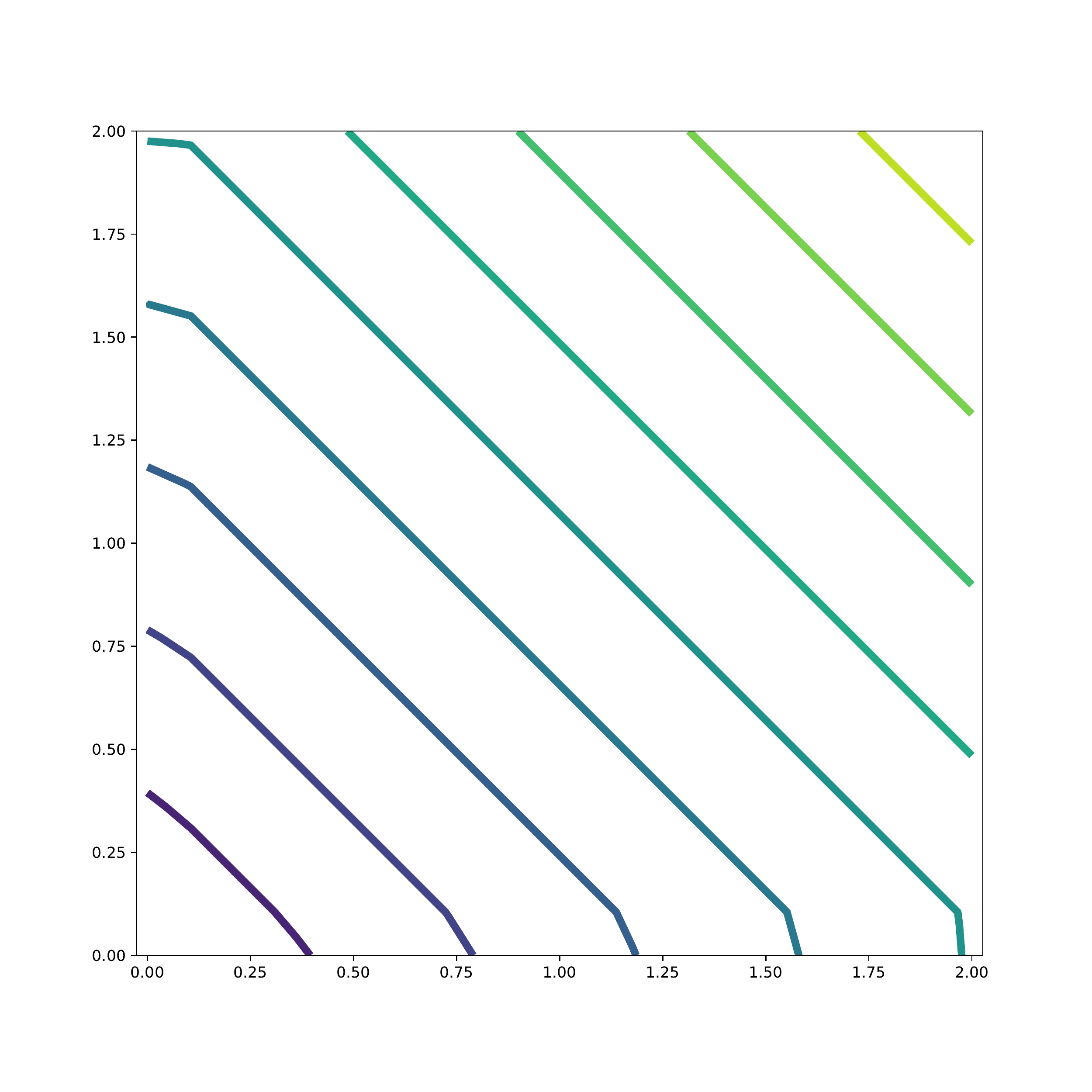}
         \caption{When there are many specialists workers in the economy, the salary becomes linear.}
         \label{fig:expes:eco:salary_specialists}
     \end{subfigure}
     \qquad
     \begin{subfigure}[b]{0.30\textwidth}
         \centering
         \includegraphics[width=\textwidth, trim = {0 1.6cm 0 1.5cm}, clip]{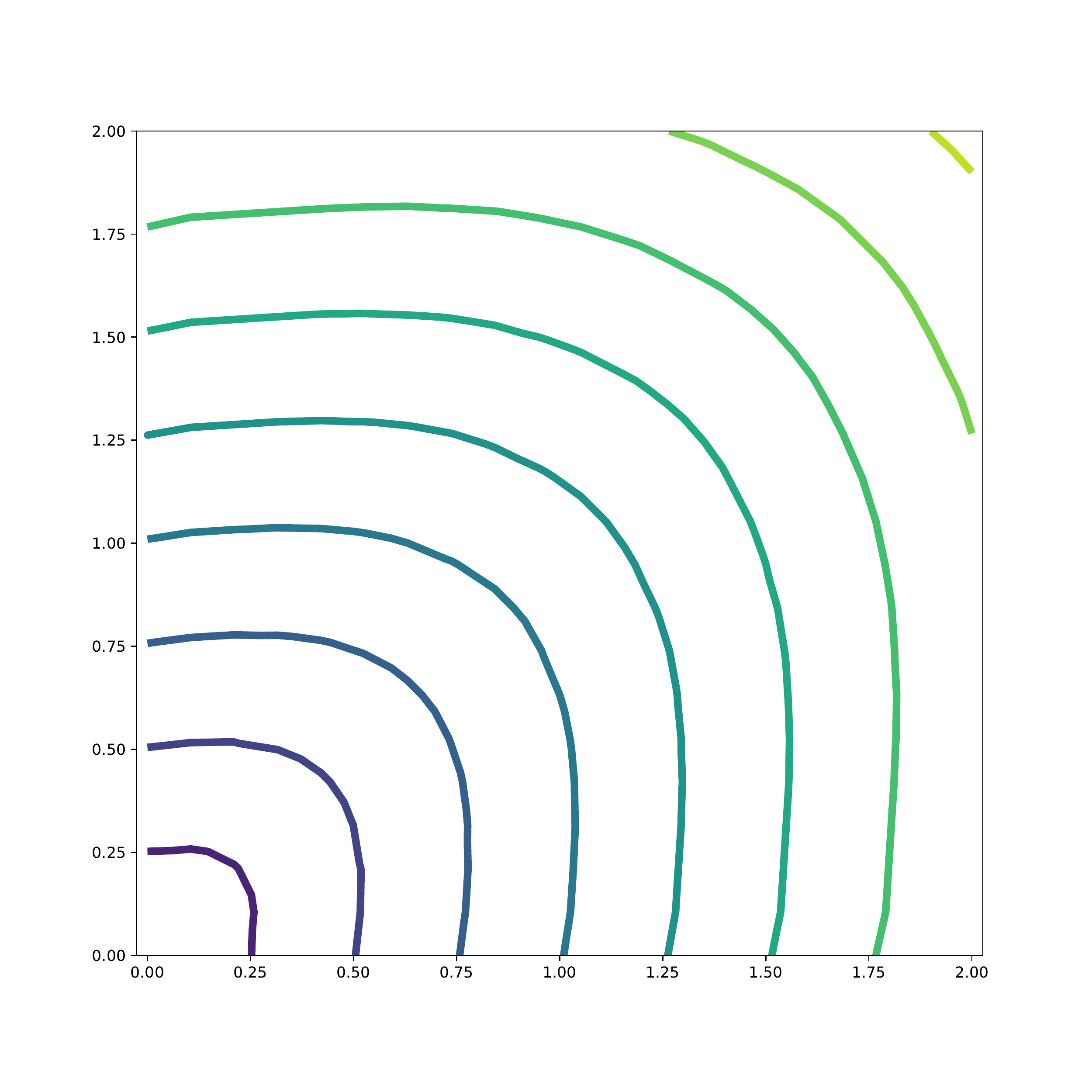}
         \caption{When there are many generalist workers in the economy, the salary is strictly convex.}
         \label{fig:expes:eco:salary_generalists}
     \end{subfigure}
        \caption{The isowage curves $\psi(x_1,x_2) = \text{constant}$ in both scenarios.}
        \label{fig:expes:eco:salaries}
    \vskip-0.5cm
\end{figure}

\looseness=-1 Figure~\ref{fig:expes:eco:alpha_to_theta} shows the aggregate skill profile of employees (the ratio of skill 2 over skill 1) as a function of the technical intensity in skill 2 of their employing firm. As predicted by~\cite{chone2021matching}, firms whose technology is very  intensive in skill~2 (i.e., high $\alpha_2$) use more skill 2 relative to skill 1 in scenario A compared to scenario B. In other words, firms are able to specialize to their ``core business'' in scenario A. This is because in that scenario (with many specialist workers), the salary tends to become linear (see Figure~\ref{fig:expes:eco:salaries})  and firms freely adjust the proportion of specialists they hire to achieve their optimal mix of skills. In contrast, in scenario B, the workers' salary is strictly convex, implying that specialist workers are expensive (compared to generalists) and hence it is too costly for firms to hire the specialists they would need to take full advantage of their technology.

\section{Conclusion and Future Work}

\looseness=-1 We have proposed the first algorithms to compute the weak optimal transport and weak optimal transport with unnormalized kernel problems in the discrete setting, both in their primal and dual formulations, as well as in the special case of barycentric cost functions. We have illustrated the interest of the WOT and WOTUK problems on the economic application of~\cite{chone2021matching}. Future work includes the statistical study of WOT ad WOTUK, as well as applications of these models in other applied sciences.

\paragraph{Acknowledgments} The three authors acknowledge support from the European Research Council (ERC), advanced grant FIRMNET (project ID 741467).

\clearpage

\bibliography{biblio.bib}
\bibliographystyle{plainnat}

\newpage
\appendix

\section{Other examples of WOT problems}
\label{app:other}
\paragraph{Entropic optimal transport}The entropic optimal transport (EOT) problem~\cite{wilson1969use} is a variant of the Kantorovich problem in which an entropic regularization term is added:
\begin{align}
    \Sinkhorn_\epsilon(\mu,\nu)
    &\defeq \inf_{\pi \in \Pi(\mu,\nu)} \iint_{\XY} c \,d\pi + \epsilon \KL(\pi | \mu\otimes\nu)
    \\
    &=
    \inf_{\pi \in \Pi(\mu,\nu)} \int c \,d\pi + \epsilon \int \log\frac{d\pi}{d\mu\otimes\nu} \,d\pi \label{eqn:wot:entropic_ot}
\end{align}
where $c \in \CXY$ is the cost function, $\epsilon > 0$ is the regularization strength, and $\KL(\cdot | \cdot)$ is the relative entropy (or Kullback-Leibler divergence). Considering the disintegration $\left(\pi_x\right)_{x \in \X}$ of $\pi$ with respect to $\mu$, and noting that $\frac{d\pi}{d\mu\otimes\nu}(x,y) = \frac{d\pi_x}{d\nu}(y)$, problem~\eqref{eqn:wot:entropic_ot} rewrites:
\[
    \inf_{\pi \in \Pi(\mu,\nu)} \int_\X \left[ \int_\Y \left( c(x,y) + \epsilon \log\frac{d\pi_x}{d\nu}(y) \right) d\pi_x(y) \right] d\mu(x)
\]
which corresponds to the WOT problem~\eqref{eqn:wot:wot} with
\begin{align*}
    \F(x,p) &= \int_\Y \left( c(x,y) + \epsilon \log\frac{dp}{d\nu}(y) \right) dp(y)
    \\
    &= \int_\Y c(x,y) \,dp(y) + \epsilon \KL(p | \nu).
\end{align*}

\paragraph{Martingale optimal transport}The martingale optimal transport (MOT) problem~\cite{beiglbock2013model} is a variant of the Kantorovich problem in which the optimal transport plan is constrained to be a martingale:
\begin{align*}
    \sup_{\substack{\pi \in \Pi(\mu,\nu)\\\mu\,\text{a.e.}, \int y \,d\pi_x(y) = x}} \iint_{\XY} F(x,y) \,d\pi(x,y).
\end{align*}
Up to the fact that $\F$ is now allowed to take value $+\infty$, this problem corresponds to the WOT problem~\eqref{eqn:wot:wot} with $\F(x,p) = \int c(x,y) \,dp(y) - \iota\left(\mu\,\text{a.e.}, \int y \,dp(y) = x\right)$ where $\iota(a)$ equals $0$ if the assertion $a$ is true, and $+\infty$ if $a$ is false.

\section{Proofs}

\subsection{Proof for the projection onto $\Pi(\ll \mu,\nu)$}
\label{appendix:proof:projection_dual_wotuk}

Projecting a matrix $P \in \Rnm_+$ onto $\Pi(\ll \mu,\nu)$ means solving the following optimization problem:
\[
    \min_{Q \in \Pi(\ll \mu,\nu)} \KL(Q | P)
    =
    \min_{\substack{Q \in \Rnm_+ \\ Q^\top\ones=b}} \KL(Q | P)
    =
    \min_{\substack{Q \in \Rnm \\ Q^\top\ones=b}} \sum_{i,j} Q_{ij} \left( \log\frac{Q_{ij}}{P_{ij}} - 1 \right)
\]
where we can drop the non-negativity constraint over $Q$ for it is already constrained by the $\log$ in the objective.

The Lagrangian of the problem is
\[
    L(Q, \lambda) = \sum_{i,j} Q_{ij} \left( \log\frac{Q_{ij}}{P_{ij}} - 1 \right) + \langle \lambda, b - Q^\top\ones \rangle
\]
where $\lambda \in \R^m$ is the Lagrange multiplier for the constraint $Q^\top\ones=b$. The problem is convex, so the first-order condition is sufficient for optimality. So the solution should verify
\[
    \log\frac{Q_{ij}}{P_{ij}} = \lambda_j
\]
hence
\[
    Q_{ij} = \frac{P_{ij} b_j}{\sum_{i'} P_{i'j}}.
\]

\subsection{Proof for the dual formulation of $\opt\psi$}
\label{appendix:proof:phi_to_psi}

Let us first compute the dual problem corresponding to
\[
    \min_{\substack{p \in \mathbb{R}_+^m \\ \sum_{j=1}^m p_j = 1 \\ \sum_{j=1}^m p_j y_j = z}} \langle p, \phi \rangle.
\]
One has:
\begin{align*}
    \min_{\substack{p \in \mathbb{R}_+^m \\ \sum_{j=1}^m p_j = 1 \\ \sum_{j=1}^m p_j y_j = z}} \langle p, \phi \rangle
    &=
    \min_{p \in \mathbb{R}_+^m} \langle p, \phi \rangle + \sup_{\lambda \in \Rq, \mu\in\R} \left\langle \lambda, z - \sum_{j=1}^m p_j y_j \right\rangle + \mu \left(1 - \sum_{j=1}^m p_j\right) \\
    &=
    \sup_{\lambda \in \Rq, \mu\in\R} \langle \lambda, z \rangle + \mu + \inf_{p \in \mathbb{R}_+^m} \sum_{j=1}^m p_j \left(\phi_j - \mu - \langle \lambda, y_j \rangle\right)\\
    &= \sup_{\substack{\lambda \in \Rq, \mu\in\R \\ \forall j, \langle \lambda, y_j \rangle + \mu \leq \phi_j}} \langle \lambda, z \rangle + \mu
\end{align*}
where we have swapped the $\min$ and the $\max$ using the strong duality theorem for linear programs.

Likewise, the dual of 
\[
    \min_{\substack{p \in \mathbb{R}_+^m \\ \sum_{j=1}^m p_j y_j = z}} \langle p, \phi \rangle
\]
will be the same as before but without $\mu \in \R$, because we have dropped the associated constraint, \textit{i.e.}
\[
    \min_{\substack{p \in \mathbb{R}_+^m \\ \sum_{j=1}^m p_j y_j = z}} \langle p, \phi \rangle
    =
    \sup_{\substack{\lambda \in \Rq \\ \forall j, \langle \lambda, y_j \rangle \leq \phi_j}} \langle \lambda, z \rangle.
\]

\end{document}